\documentclass{esannV2}
\usepackage[dvips]{graphicx}
\usepackage[utf8]{inputenc}
\usepackage{amssymb,amsmath,array}

\usepackage{hyperref}

\usepackage{dirtytalk}
\usepackage{xcolor}
\newcommand*{\mybox}[2]{\colorbox{#1!30}{\parbox{\linewidth}{#2}}}
\usepackage{scalerel}

\begin{document}
\title{Beyond Supervised Continual Learning: a Review}

\author{%
Benedikt Bagus$^1$, Alexander Gepperth$^1$ and Timothée Lesort$^2$
%
\vspace{.3cm}\\
\scriptsize $^1$ University of Applied Sciences Fulda - Department of Computer Science\\
\scriptsize Leipzigerstraße 123, 36037 Fulda - Germany
\vspace{.1cm}\\
\scriptsize $^2$ Montréal University (UdeM), Mila\\
\scriptsize 6666 St-Urbain Street, Montréal, QC - Canada
}

\maketitle
\begin{abstract} 
  \textit{Continual Learning} (CL, sometimes also termed \textit{incremental learning}) is a flavor of machine learning where the usual assumption of stationary data distribution is relaxed or omitted.
  When naively applying, e.g., DNNs in CL problems, changes in the data distribution can cause the so-called \textit{catastrophic forgetting} (CF) effect: an abrupt loss of previous knowledge.
  Although many significant contributions to enabling CL have been made in recent years, most works address supervised (classification) problems.
  This article reviews literature that study CL in other settings, such as learning with reduced supervision, fully unsupervised learning, and reinforcement learning.
  Besides proposing a simple schema for classifying CL approaches w.r.t. their level of autonomy and supervision, we discuss the specific challenges associated with each setting and the potential contributions to the field of CL in general.
\end{abstract}

\section{Introduction}
\label{sec:intro}
Continual learning is a field of machine learning where the data distribution is not static.
It is a natural framework for many practical problems where the data arrives progressively, and the model learns continuously.
For example, in robotics, robots need to adapt to their environments to interact and realize actions constantly, or recommendation systems also need to adapt constantly to the new content available and the new needs of users.
However, in recent years, the field of continual learning has focused mainly on one type of classification scenario: class-incremental.
This scenario evaluates how models can learn a class once and remember it when new class data arrives.
While it is important to solve this problem, using only one type of scenario can lead to over-specialized solutions that cannot generalize to different settings.
In this paper, we propose to review the literature dealing with other settings than the default one (class-incremental) and, more generally, fully supervised scenarios.
The goal is to shed light on efforts made to diversify the evaluation of continual learning.

We introduce the continual learning framework and the goals of continual learning (Sec~\ref{sec:framework}).
Then, we describe the default scenario and its characteristics (Sec~\ref{sec:default}).
In addition, we introduce a scenario that goes beyond the default scenario in supervised learning (Sec~\ref{sec:generalization_default}), unsupervised learning (Sec~\ref{sec:unsupervised}) and reinforcement learning (Sec~\ref{sec:reinforcement}).
\\
\mybox{gray}{
  \textbf{Disclaimer: } This article compares the differences between supervised continual learning (CL) and other settings.
  Each of these settings can have appropriate use cases and application fields.
  Therefore, the goal is not to push for a different kind of CL that is supposedly more \say{natural} or \say{realistic}, but to point out that other feasible settings for CL exist, with partially overlapping challenges and solutions.
  Thus, we review existing literature, list commonly made assumptions, and point out remaining challenges specific to non-supervised continual learning.
  Moreover, benchmarking diversity is of high value if different benchmarks are built with the intent to evaluate one particular criterion (of which there are several).
  Benchmarks or scenarios that are not built for such purposes may contribute less to progress in CL.
}

\section{Framework and goals of continual learning}
\label{sec:framework}
Continual learning (CL) is a machine learning sub-field that studies learning under time-varying data distributions.
This relaxes one of the fundamental assumptions of statistical learning theory \cite{vapnik1999nature}, which states that the data follows a stationary distribution.
One advantage of this assumption is its simplicity, whereas CL scenarios are very diverse, depending on the nature of the non-stationarity.
In CL, it is therefore crucial to clearly define the scenario, the goals of learning, the evaluation measures, and the loss functions.
The following section describes typical non-stationarities of the data distribution that have been considered in the literature (see also \cite{gepperth2016incremental,lesort2021understanding}).

\subsection{Data distribution drifts}
\label{sub:drifts}
CL under data distribution shifts needs memorization mechanisms adapted for the type of non-stationarity, which requires assumptions by the used algorithms in the face of an infinite number of possible ways a distribution can be non-stationary.
For simplicity, we will list several typical definitions used for supervised learning (e.g., classification) but that can be generalized to other settings.

A simple way to categorize non-stationarities is based on class information.
We may distinguish two types of shifts \cite{gepperth2016incremental} in this case: \textit{concept shift}, where the annotation of existing data changes \cite{caccia2020online} and \textit{virtual shift}, where we get new data, but the annotation does not change.
Usually, the term \textit{shift} is employed for sudden changes in data distributions, whereas the term \textit{drift} is used for gradual changes over time.
In supervised CL, virtual shifts are the most common non-stationarities that have been studied.
We can distinguish the special cases of \textit{virtual concept shift}, implying new data and new labels, and \textit{domain shift}, where new data of known labels are observed.
Those two settings are also known as respectively \textit{class-incremental} and \textit{instance-incremental} \cite{Lesort2019Continual,van2019three}.

The objectives to be optimized may change over time as well \cite{lesort2021understanding}, as in continual reinforcement learning \cite{Normandin2021,Khetarpal2020,Bagus2022}.

\subsection{Common CL constraints}
\label{sub:constraints}
If CL were not subject to constraints, there would be a simple solution to any scenario.
It involves storing all incoming samples and re-training every time a decision is expected, \cite{prabhu2020gdumb}.
This entails a time and memory complexity that is at least linear in the number of processed samples.

However, most CL proposals assume that memory is limited in some way, preventing this (obvious) solution.
Many approaches similarly assume that the storage of samples is restricted.
Other resources constraints subject of study are: \textit{computational cost}, \textit{memory}, \textit{data privacy}, \textit{fast adaptation}, \textit{inference speed}, \textit{transfer}.
Other constraints that are more related to the reliability of approaches are \textit{stability} and \textit{explainability}.
A discussion of CL constraints can be found in, e.g., \cite{pfuelb2019}, whereas evaluation measures that take these constraints into account are given in \cite{Lesort2019Continual}.

\section{The default scenario for CL}\label{sec:default}
\begin{figure}
  \centering
  \includegraphics[width=\textwidth]{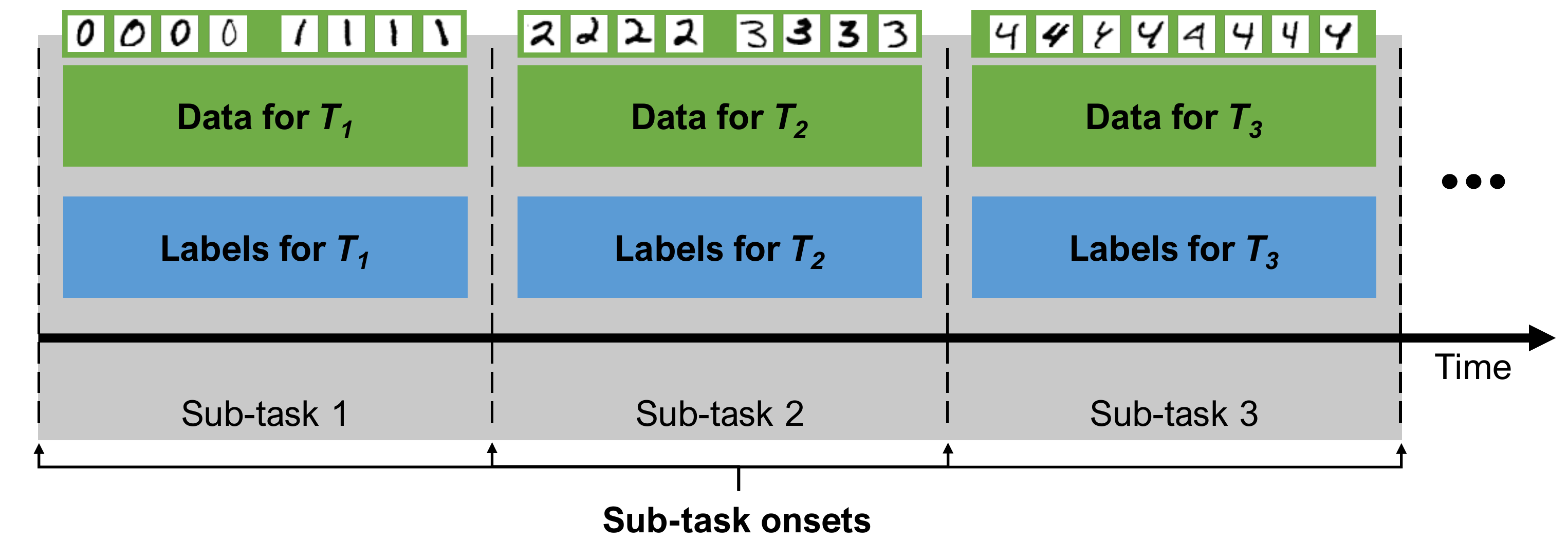}
  \caption{
    The default scenario for CL.
    The data stream is assumed to be partitioned into \textit{sub-tasks} defined by data and labels (targets).
    Data statistics within a sub-task are assumed to be stationary.
    In addition, sub-task data and labels are assumed to be disjoint, i.e., from different classes.
    The sub-task onsets are generally assumed to be known as well.
  }
  \label{fig:cl:default}
\end{figure}
A particular supervised setting, which we will refer to as the \textit{default scenario}, is currently dominating CL literature.
It is based on a classification problem divided into a small number of \textit{sub-tasks}.
Virtual concept shift occurs abruptly at sub-task boundaries by the apparition of new samples belonging to previously unseen classes, see Fig~\ref{fig:cl:default}.
Usually, the sub-task onsets/boundaries are known.
A consequence of this disjointness of sub-tasks is that no pure concept shift is involved: the annotation for a given data point will never change or be subject to conflict as learning progresses.

In the default scenario, the goal of CL is usually to learn the complete statistics of all sub-tasks as if they had been processed all at once, rather than one after the other.

Sometimes, samples are processed one by one, or all samples in a given sub-task are simultaneously available.
In some works, the sub-task index is known at test time, which is used for selecting the correct head of a multi-headed DNN for inference.

The assumptions made in the default scenario are justified in many use cases.
However, it is obvious that other scenarios, e.g., in robotics, may be found where they impose too severe restrictions.
Moreover, many characteristics of the default scenario, such as \say{drift are abrupt} or, \say{tasks are not revisited} lend themselves to \textit{benchmark overfitting}.
As an example, consider sub-task $T_2$ in Fig~\ref{fig:cl:default}: here, a DNN could punish incorrect decisions for class \say{$1$} more strongly than incorrect decisions for classes from the current sub-task since the default scenario assumes that sub-tasks are disjoint.
Even if researchers do not consciously exploit these assumptions, the employed CL algorithms may still rely on them indirectly.
It is thus fundamental to perform experiments in scenarios where these assumptions do not hold.

Thus, creating diverse benchmarks, as well as approaches that do not critically rely on the assumptions from the default scenario, should be an ongoing effort.
This effort should be pushed notably by existing continual learning libraries such as Continuum \cite{douillard2021continuum}, Avalanche \cite{continualai2021avalanche} or Sequoia \cite{Normandin2021}.

\subsection{CL approaches for the default scenario}
\label{sub:approaches}
This section is not meant to be exhaustive, since the default scenario is not the focus of this article.
Please refer to recent reviews \cite{deLange2019continual,BELOUADAH202138} for more details on CL methods for the default scenario.

Broad strategies for performing CL in the default scenario are regularization \cite{kirkpatrick2017overcoming,lopez2017gradient}, replay \cite{rebuffi2017icarl, Rolnick2019,shin2017continual} and dynamic architectures \cite{Rusu16progressive,fernando2017pathnet,veniat2021efficient,ostapenko2021continual,mendez2022modular}.
Regularization penalizes changes to model parameters that are deemed important for past sub-tasks.
This is usually achieved by adding penalty terms to the loss function, and it is implicitly assumed that new sub-tasks add only new data and classes.
Dynamic architecture methods extend models over time in order to separate previously learned parameters from currently optimized ones, thus reducing cross-talk and catastrophic forgetting (but equally assuming that new sub-tasks contain only new data).
Replay methods store received data for subsequent use in re-training (rehearsal).
Instead of relying on stored data, re-training can also be performed using samples produced by generative models (generative replay).

Replay is known to be an effective method for preventing catastrophic forgetting (CF), especially in class-incremental settings \cite{lesort2021continual,deLange2019continual}, but also for continual reinforcement learning \cite{Kalifou2019,Traore2019,Rolnick2019} or unsupervised learning \cite{lesort2018generative,rao2019continual,madaan2021representational}.

\subsection{Metrics and evaluation procedures}
\label{sub:metrics}
In the default scenario, various measures related to the classification error are common, which have been discussed in, e.g., \cite{Kemker2017Measuring,diaz2018don,maltoni2019continuous}.
A common baseline is termed \textit{cumulative performance}, obtained by evaluating models on the merged test sets from all sub-tasks, which corresponds to learning with stationary statistics.
This baseline is often considered an upper bound for CL performance.

In addition, \cite{lopez2017gradient} proposed the notions of forward and backward transfer: forward transfer (FT) measures how training on sub-task $i$ impacts performance on a future sub-task $j>i$.
For backward transfer (BT), the impact on previous sub-tasks $j<i$ is considered.
The common case in CL is negative BT indicating forgetting, but positive BT is theoretically possible as well.

Many authors, e.g., \cite{kirkpatrick2017overcoming}, assume that (although sub-tasks are presented sequentially) all sub-task data are available for model selection and hyper-parameter tuning.
For example, tuning EWC's regularization strengths $\lambda_i$ for each sub-task is often done in hindsight.

Some authors \cite{serra2018overcoming,Doan2021Theoretical,farajtabar2019orthogonal}, especially in works using multi-head DNNs, assume that the sub-task ID is known during testing, although this does not seem to be the current consensus.
In the limit where each sub-task contains only a single class, providing the sub-task ID at test time means providing the class label.
Even if sub-tasks are more diverse, the sub-task ID contains significant information and may thus confer unfair advantages.
The question of evaluation protocols in CL is discussed in \cite{pfuelb2019,pfuelb22diss}.

\subsection{Benchmarks} 
\label{sub:benchmark}
Benchmarks for the default CL scenario are mostly derived from datasets such as MNIST, CIFAR10/100, Imagenet, SVHN etc. to create \textit{class-incremental} or \textit{domain-incremental} scenarios.
The \textit{permuted MNIST} benchmark, where successive sub-tasks are created by permuting all pixels according to a sub-task specific permutation scheme, was initially popular \cite{kirkpatrick2017overcoming,lopezpaz2017gem,Wortsman2020Supermasks} but is less so now because it can, to good accuracy, be solved even without dedicated CL schemes \cite{pfuelb2019}.
Some authors used Atari games \cite{mnih2013atari}, Mujoco \cite{Todorov2012mojoco} or Meta-World \cite{Yu2019} as benchmarks.
CL specific variants of standard benchmarks such as, e.g., \textit{colored MNIST} \cite{kim2019learning} are widely used as well since they can be used to investigate specific aspects of CL, see \cite{gat2020removing,li2019repair}.

\section{Generalizations of the default scenario}
\label{sec:generalization_default}
The default scenario is convenient for evaluation and represents a rather controlled setting for CL.
In less controlled settings, fully annotated data may not be available, or supplementary constraints may be imposed.
We find it convenient to introduce a new taxonomy of CL approaches based on their level of autonomy.

\subsection{Classifying the autonomy level of CL algorithms}
\label{sub:autonomy}
The various applications of continual learning can be classified in autonomy levels, as for autonomous vehicles \cite{kiran2021deep}.
Obviously, CL should get harder as less and less human supervision is supplied.
We identify two dimensions of autonomy, both of which will be discussed in-depth to characterize generalized supervised CL approaches better.

\par\smallskip\noindent\textbf{Objective Autonomy }
denotes the autonomy regarding the objective to achieve (labels, targets, rewards), which we group into 4 levels:
\begin{itemize}
  \item \textbf{Level 0:} Full data annotation: supervised.
  \item \textbf{Level 1:} Sparse labelization: RL, active learning, sparse training.
  \item \textbf{Level 2:} No annotation for training, query for fast adaptation.
  \item \textbf{Level 3:} No annotation for training, zero-shot adaptation.
\end{itemize}
For objective supervision levels 2 and 3, continual training can be seen as pretraining for the unknown future objective task.
We can note that if the scenario objective is unsupervised, then we can assume that it is similar to full data annotation.

\par\smallskip\noindent\textbf{Continual Learning Autonomy }
is concerned with autonomy regarding the distribution shifts (task label, task boundaries):
\begin{itemize}
  \item \textbf{Level 0:} Full task annotation at train and test.
  \item \textbf{Level 1:} Full task annotation at train, no test task labels.
  \item \textbf{Level 2:} Sparse task annotation at train, no test task labels: example task boundaries only without train task label.
  \item \textbf{Level 3:} No task labels at all: task-agnostic.
\end{itemize}

This classification still holds for smooth transitions (concept drift).
We can note that for class-incremental problems, the task-agnostic setting does not make sense since the task information is in the class labels.
We can now characterize CL contributions by a pair \say{($i$~$j$)}, where $i$ represents the objective autonomy level, and $j$ stands for the CL autonomy level.
Hence, the pair (0~1) describes a class-incremental scenario, i.e., a classification setting with fully annotated data and task labels for training but not for test data.
It is important to note that those ratings classify the complexity of a given scenario and not approaches.
For example, the default scenario (class-incremental) assesses a complexity level of (0~1), domain incremental without task labels would be (0~2), task agnostic continual reinforcement learning \cite{Caccia2022} (1~3).
To validate approaches' autonomy, they should be evaluated on adequate scenarios.

Investigating the cost of lowering or increasing a task's complexity level is fundamental for applications of CL.
We want to make our algorithms scale up to arbitrary complexity levels, but in practice, we would always choose the lowest possible complexity.
If permitted by a given application, solving a scenario with a complexity level (0~1) is obviously more efficient than solving the same scenario with a level of (3~3).

\subsection{Towards a generalization of the default setting}
Some variants of supervised CL exist that alleviate the need for annotated data.
The reduction of annotations can apply to restricted access to the task labels as in task-agnostic CL \cite{zeno2018task,He2019TaskAC}, or by reducing the labels' availability as in continual active learning \cite{mundt2020wholistic,perkonigg2021continual}, or in semi-supervised continual learning, \cite{smith2021memory}.
As described in Sec~\ref{sub:autonomy}, reducing access to task supervision lead to evaluating a CL autonomy level of 2, and removing all access to it lead to evaluating a CL autonomy level of 3.
On the other hand, reducing data annotation lead to an objective autonomy of 1 instead of 0 when the full annotation is available.

Among potential supplementary constraints, data can be streamed without the possibility of multi-epoch training as in online training \cite{Chaudhry19,schwarz2018progress}, or the data can be imbalanced \cite{kimimbalanced,chrysakis2020online}, or mixed with spurious features \cite{lesort2022Spurious}.
Scenarios where the annotations change over time (real concept drift) have been investigated in \cite{lesort2021understanding,abdelsalam2021iirc,caccia2020online}.
Some contributions \cite{dz22} relax the condition of disjoint sub-tasks and assess the impact of several fundamental CL strategies such as regularization and replay.
Yet others \cite{pfuelb2021a,pfuelb2021b} demonstrate that detecting sub-task boundaries autonomously is generally feasible by using density estimation methods.

To the generalized settings cited above, supplementary constraints (as discussed in Sec~\ref{sub:constraints}) may be added, making them even harder.
If this brings CL closer to real-life applications, solutions are required that do not overfit a particular CL setting.
Currently, the field of CL is fundamentally meta: implicitly, the goal is not to train the best possible model w.r.t. non-continual baselines, but rather to create algorithms that show maximal generalization to other CL settings.
Therefore, experimenting with generalized supervised scenarios can assess algorithms' robustness and improve generalized CL.

\section{Unsupervised continual learning}
\label{sec:unsupervised}
Whether a machine learning task is considered supervised or not depends on the formulation of the loss function.
In fact, no assumptions whatsoever are made concerning the loss in the definition of CL given in Sec~\ref{sec:intro}.
Therefore, CL is naturally transferred to unsupervised methods of machine learning, typical examples of which are density modeling, clustering, generative learning, and unsupervised representation learning.

\subsection{Density modeling}
Density modeling aims at approximating the probability density of a given set of data samples directly by minimizing a log-likelihood loss.
Typically, this is achieved using \textit{mixture models}, which model the data density as a weighted sum of $N$ parameterized component densities, e.g., multivariate Gaussian densities or Dirichlet distributions.
Density modeling allows performing, among other functions, Bayesian inference and sampling.
These functionalities spawned increased interest in mixture modeling a few years back, particularly in robotics \cite{pinto2015fast,shmelkov2017incremental,pokrajac2007incremental,kristan2008incremental}.
The main issue for CL in mixture modeling is that concept drift or shift may require an adaptation of $N$, which motivates heuristics for adding and removing components.
Current approaches using generative replay are proposed in, e.g., \cite{rao2019continual}.
Mixture models usually adapt only a small subset of components for each update step due to their intrinsic reliance on distances instead of scalar products.
This is why they are less prone to catastrophic forgetting than DNNs, an effect that has been demonstrated for self-organizing maps in \cite{gepperth2020a} which are an approximation to Gaussian Mixture Models (GMMs), see \cite{gepperth2019}.
Modeling the data density allows partitioning data space into Voronoi cells, in each of which a separate linear classifier model can be trained.
This is the essence of the popular Locally Weighted Projection Regression (LWPR) algorithm \cite{vijayakumar2000locally} which was explicitly constructed for continual classification in robotics.

\subsection{Clustering}
Clustering is, in a certain sense, an approximation to density modeling, although the inference is limited to determining the precise component a given data sample was generated from.
Clustering methods are normally trained using a k-means type of algorithm, which approximates gradient descent on a loss function that again approximates a GMM log-likelihood.
CL for clustering algorithms faces the same basic issue as in density modeling: a potentially variable number $N$ of cluster centers during concept drift or shift.
This has been demonstrated in, e.g., \cite{pham2004incremental,aaron2014dynamic,bagirov2011fast}.

\subsection{Generative learning}
Generative learning aims to generate realistic samples (typically images) that are similar to a set of training data.
Typical models are generative adversarial networks (GANs), variational auto-encoders (VAEs), PixelCNN, FLoW or GLoW, but many other variants have been proposed, see, e.g., \cite{turhan2018recent} for a review.
Training such generative models can be performed, e.g., in the CL default scenario introduced in Sec~\ref{sec:default} (apart from the supervision information), which leads to catastrophic forgetting (CF) without additional measures.
Several of the approaches used in supervised learning have been successfully applied to training generative models: knowledge distillation \cite{gancl1}, EWC \cite{zhai2019lifelong} and replay \cite{8682702,9190980,lesort2018marginal,ramapuram2017lifelong,lesort2018generative}.
To our knowledge, no generic approaches that are specific to generative learning have been proposed, apart perhaps \cite{varshney2021cam} where it is proposed to learn specific transformations.
This, however, is very specific to a particular kind of (image) data and would have to be adapted if other kinds of data were targeted.

\subsection{Continual representation learning}
\label{sub:representation}
Unsupervised training for learning representations for downstream applications is a common use case for unsupervised learning.
It was one of the motivations to develop various types of auto-encoders and generative models in the early days of deep learning.
In CL, using an unsupervised criterion to learn representations might be useful to avoid representations that overfit a specific task and, at the same time, improve performance on downstream tasks \cite{fini2022self,rao2019continual,madaan2021representational}.
Unsupervised pre-training can also be useful for learning a general feature extractor that can be frozen for future tasks \cite{Traore19DisCoRL,Ostapenko2022Foundational,caccia2021special}.

\subsection{Challenges of unsupervised CL}
Unsupervised learning offers general learning criteria that can avoid the over-specialization of supervised training and reduce forgetting.
Nevertheless, unsupervised CL faces the same challenges as supervised CL, and the default scenario for supervised CL can be transferred.
Moreover, in practice, unsupervised training tends to be more complex than supervised training, especially for generation and density modeling, since it is harder to model a distribution than to determine a separating hypersurface between classes in data space.
With the added complexity of CL, unsupervised learning can be a formidable problem, especially w.r.t. model and hyper-parameter selection.

\section{Continual reinforcement learning}
\label{sec:reinforcement}
In reinforcement learning (RL), an agent learns to interact with its environment by choosing a specific action for each state based on a reward signal.
The (unknown) underlying process is formalized as a Markov decision process (MDP), where an optimal policy maximizes an expected reward.
This scenario is inherently a CL setting, since the distribution of the observed data depends on the specific policy.
The evolution of the policy throughout the learning process will mechanically lead to the non-stationarity of the data distribution.
Hence, RL requires the ability to cope with non-stationary data.
However, supplementary non-stationary, for example, in the environment or in the objective to fulfill, can increase the training difficulty and lead to a continual setting.
We will use the term \textit{Continual Reinforcement Learning} (CRL) for denoting RL in settings that go beyond the usual assumptions of non-stationarity made in conventional RL.

\subsection{Existing approaches}
\label{sub:RW_RL}
The works presented in \cite{ring94,Thrun1995} introduce the importance of CL at an early stage and especially investigated them in the context of reinforcement learning.
More recent works, e.g., \cite{Xu2018,Rolnick2019} revisit this area and consider additional aspects such as catastrophic forgetting.
Some frameworks to guide future research have also been published \cite{Lesort2019,Khetarpal2020}.
Both provide a comprehensive overview of the synergies between continual and reinforcement learning.

\par\smallskip\noindent\textbf{RL Approaches }
Experience replay \cite{Zhang2017,Rolnick2019,Fedus2020} is the most common approach to counter non-stationarities in RL.
Several variants are introduced, e.g., \cite{Schaul2015,Andrychowicz2017,Isele2018,Novati2019,Hu2021}.

Continuous control, multi-task, and multi-goal are also research topics intersecting with continual reinforcement learning, but their scenarios are not always defined in a consistent fashion in the literature.
In general, the goal is to enable transfer learning between policies, which, however, omits the capacity for forgetting or re-adaptation.
Some works assume a static objective \cite{Ammar2014,Teh2017,Sorokin2019,Ribeiro2019,Schiewer2021}, others a static agent and/or environment \cite{Zhao2019,Yang2020,Gupta2021} or none of both \cite{Xu2020a,Kalashnikov2021,Kelly2021}.
In contrast, multi-agent reinforcement learning is mostly related to some kind of joint training and is hence not related to CL.

For CRL, the agent needs to acquire new skills to handle time-varying conditions, such as changes in environment \cite{Padakandla2021}, observations or actions, and additionally must retain the old knowledge.
A variety of approaches has been published, among which knowledge-based distillations \cite{Kalifou2019,Traore2019} and context-based decompositions \cite{Mendez2021,Zhang2022} are popular.
Other works are concerned with the employed model \cite{Kaplanis2018,Kaplanis2019,Lo2019,Huang2021}, off-policy algorithms \cite{Xie2020}, policy gradient \cite{Mendez2020} or a task-agnostic perspective \cite{Caccia2022}.
Evaluations of known CL methods (e.g., GEM, A-GEM, and replay) are also applied in the RL domain \cite{Atkinson2021,Bagus2022}.

\par\smallskip\noindent\textbf{Benchmarks }
An overview of CRL environments can be found in \cite{Khetarpal2018}.
Dedicated benchmarks which allow a systematical assessment are: \textit{Meta-World} \cite{Yu2019}, \textit{Continual World} \cite{Wolczyk2021} and \textit{L2Explorer} \cite{Johnson2022}.

\par\smallskip\noindent\textbf{Libraries }
Some libraries aim at unifying CRL development to improve comparability and accelerate progress:
\textit{Sequoia} \cite{Normandin2021}, \textit{Avalanche rl} \cite{Lucchesi2022}, \textit{SaLinA} \cite{denoyer2021salina}, \textit{Reverb} \cite{Cassirer2021} and \textit{CORA} \cite{Powers2021}.

\subsection{Assumptions in CRL}
Three assumptions are commonly made in CRL:
Foremost, a decomposition into sub-tasks is assumed, even if their onset is unknown since most dedicated CL methods (see Sec~\ref{sub:approaches}) assume the existence of distinct sub-tasks.
Another assumption concerns samples, which are assumed to be non-contradictory within sub-tasks, meaning the assessment of rewards changes only between sub-tasks, if it changes at all.
Finally, it is a common assumption that knowledge of sub-task boundaries is provided.
Most existing works are using information about sub-task boundaries as if they were provided by an oracle, without the possibility to recognize or determine them autonomously.

\subsection{Challenges}
In CRL, various types of drifts/shifts can appear:
\par\smallskip\noindent\textbf{Environment-related } 
The agent successively observes its environment.
Therefore, on a short timescale, observations will always be non-stationary, even if the environment is.
In addition, the environment itself can change over time, or rapid modifications can be encountered (\textit{environment shift}).
This would result in novel states or transitions between these, resulting in an enlargement of the actual involved MDP.

\par\smallskip\noindent\textbf{Goal-related } 
By maximizing the reward signal, the agent attains a defined objective.
If the reward function changes, the agent experiences divergent information, leading to an inconsistent policy.
In this setting, the definition of states, actions and transitions does not change, so the underlying MDP remains structurally intact.
However, other rewards are assigned to previously learned mappings, enforcing changes of transition probabilities.

\par\smallskip\noindent\textbf{Agent-related } 
The decreasing influence of exploration, regardless of whether off-policy methods such as Q-Learning or on-policy methods such as policy gradient are used, temporarily creates a source of non-stationarity, resulting in a time-varying sampling of the state-action space even with a static policy and a static environment.
Additionally, it is easily possible that sensors or actuators degrade or undergo deliberate manipulations.
Affecting possible actions, immediate effects on the MDP, while a changed perception of states also impacts transitions.

\par\smallskip\noindent\textbf{Sub-tasks and data acquisition }
For scenarios where the environment changes in a discrete fashion, we can introduce the notion of sub-tasks as in the default scenario for supervised CL, see Sec~\ref{sec:default}.
A general challenge stems from the fact that samples are acquired as an online time series and have no balancing guarantees at all.
Moreover, it is possible that similar states and actions appear in various sub-tasks, but with different assigned rewards, so sub-tasks are usually not disjoint and may even be contradictory, requiring un- or re-learning, a concept absent from the default supervised CL scenario.
Depending on the type of non-stationarity, sub-task onset can be unknown, and the detection of boundaries may be difficult if the drifts are gradual rather than abrupt.
In addition, the number of sub-tasks can be significantly higher than in supervised scenarios, up to a point where the entire concept of sub-tasks becomes questionable.
Lastly, actions must be explicitly performed to transition to the appropriate subsequent state.
Therefore, a generative or offline sampling is of limited usefulness, at least for exploration.

\section{Discussion}
The field of CL has expanded rapidly in recent years, which is why many aspects of CL are still fluid and not subject to a common consensus among researchers.
This is evidenced by a wide variety of assumptions, evaluation metrics, see Sec~\ref{sub:metrics} and constraints, see Sec~\ref{sub:constraints}.
The so-called default scenario, see Sec~\ref{sec:default}, is the nearest thing to a commonly agreed scenario, yet many details fluctuate strongly between contributions.
This leads to several interesting consequences and opportunities for further research:

\par\smallskip\noindent\textbf{CL comparability }
A direct consequence is the difficulty to directly compare results of different articles.
This underscores the need, in CL more than in other domains of machine learning, to precisely describe evaluation procedures and, where possible, make use of existing libraries (see Sec~\ref{sec:default} and \ref{sub:RW_RL}) and evaluation procedures.
Furthermore, as stated in Sec~\ref{sub:constraints}, CL is a multi-objective problem where achieving the cumulative baseline is important, but where other measures (see Sec~\ref{sub:metrics}) matter as well.

\par\smallskip\noindent\textbf{CL autonomy }
As explained in Sec~\ref{sub:autonomy}, CL approaches should also be evaluated based on the complexity and autonomy of the scenario they can generalize to, to prevent them from overfitting to a specific CL scenario or assumptions.

\par\smallskip\noindent\textbf{CL scalability }
An aspect that is often omitted in current works in favor of quantitative performance measures is scalability.
Depending on a potential application context, CL, even in the default scenario, may be faced with a huge number of sub-tasks, each again containing enormous amounts of samples.
If this were not the case, the cumulative baseline, or equivalently some variant of GDumb (see Sec~\ref{sub:metrics}), would be a much less costly and superior (w.r.t. performance) alternative to using dedicated CL methods.
So time and memory complexity for the case where the number of sub-tasks is large should be included in all new works on CL algorithms to ensure comparability, at least in this respect.

\par\smallskip\noindent\textbf{CL generalization }
As was shown in Sec~\ref{sec:generalization_default}, \ref{sec:unsupervised} and \ref{sec:reinforcement}, the default CL scenario of Sec~\ref{sec:default} can be generalized in many ways.
Moreover, these chapters show that many open issues remain, both technical and conceptual, when attempting to generalize CL.

\section{Conclusion}
This review article attempts to give an overview of the current state of CL beyond the purely supervised default scenario, see Sec~\ref{sec:default}.
We describe the various complexification of the default scenario and the different learning paradigms, and propose a classification based on the autonomy characteristics of algorithms.
We believe that attempts to generalize CL pose important questions about the fundamental assumptions behind CL.
We thus encourage CL researchers to carefully reflect upon the implicit, hidden assumptions in each CL approach they are dealing with and whether they can (and should) be relaxed.
In a still-fluid field such as CL, a continuous re-examination of assumptions may lead to new solutions that strongly contribute to the advancement of the field.


\begin{footnotesize}
\bibliographystyle{unsrt}
\bibliography{continual_full,continual_RL,others}
\end{footnotesize}


\end{document}